\documentclass[conference]{IEEEtran}
\IEEEoverridecommandlockouts
\usepackage{cite}
\usepackage{amsmath,amssymb,amsfonts}
\usepackage{dsfont}
\usepackage{algorithmic}
\usepackage{graphicx}
\usepackage{textcomp}
\usepackage{xcolor}
\usepackage{colortbl}
\usepackage{booktabs}
\usepackage{enumitem}
\usepackage{algorithm}
\usepackage{algorithmic}
\usepackage{amsmath}
\usepackage{amssymb}
\usepackage[final]{microtype}
\DeclareFontShape{OT1}{ptm}{m}{scit}{<-> ssub * ptm/m/sc}{}
\def\BibTeX{{\rm B\kern-.05em{\sc i\kern-.025em b}\kern-.08em
    T\kern-.1667em\lower.7ex\hbox{E}\kern-.125emX}}
\begin{document}

\title{TRACE: State-Aware Query Processing over Temporal Evidence Graphs for Conversational Data
}

\author{
\IEEEauthorblockN{Maolin Wang\textsuperscript{1}, Yu Wang\textsuperscript{1}, Zichun Liu\textsuperscript{1}, Baiyuan Qiu\textsuperscript{2}, \\  Chenbin Zhang~\textsuperscript{2}, Jiguang Shen\textsuperscript{2}, Haoran Yang\textsuperscript{3}, Hao Miao\textsuperscript{4}}
\IEEEauthorblockA{\textsuperscript{1}Hong Kong Institute of AI for Science, City University of Hong Kong, Hong Kong SAR, China\\
\{morin.wang@, wangyu6-c@my.\}cityu.edu.hk}
\IEEEauthorblockA{\textsuperscript{2}Independent Researcher, Beijing, China}
\IEEEauthorblockA{\textsuperscript{3}Central South University, Changsha, China}
\IEEEauthorblockA{\textsuperscript{4}The Hong Kong Polytechnic University, Hong Kong SAR, China}
% Corresponding Author: yhr.cse@csu.edu.cn
}
\maketitle

\begin{abstract}
Conversational data is increasingly used as a persistent source of user state
for long-running assistants and AI agents. However, querying this data remains challenging
because conversations naturally evolve: plans are revised, preferences change, and later
messages frequently supersede or contradict earlier information. Existing long-memory
pipelines largely treat memories as independent text or vector objects. This approach often retrieves semantically similar but stale evidence, offering limited
support for state-aware reasoning. To address this problem, we present TRACE, a query processing framework over temporal evidence graphs for evolving conversational data.
TRACE models conversations as a hierarchical graph spanning events, sessions, and
topics, enriched with typed temporal, causal, update, and contradiction
relations. Crucially, the framework maintains validity annotations so obsolete facts remain accessible for historical queries but are discounted for current-state
answers. At query time, TRACE combines vector-based note retrieval with
graph-guided evidence search, generating validity-aware support paths and a
hybrid context for answer generation. This design separates lexical recall
from evidence reconstruction, enabling bounded query-time reasoning over
long conversational histories. Experiments on long-conversation query-answering~(QA) benchmarks
show that TRACE improves temporal and multi-hop reasoning, with ablations
highlighting the importance of hierarchy, update-aware seeding, and
path-grounded evidence. Source code is provided at https://github.com/MorinWang/TRACE.
\end{abstract}


\section{Introduction}
\label{sec:introduction}

Conversational data is now a permanent fixture in personal assistants,
enterprise copilots, customer service systems, and autonomous AI agents~\cite{zhong2024memorybank,xu2022beyond,maharana2024evaluating,wu2024longmemeval}.
Unlike static documents, this data captures shifting user states: preferences
evolve, plans change, commitments expire, and new messages often update or
contradict older ones. Querying such data requires more than retrieving relevant
text spans: a system must identify which facts are still valid, which are
outdated history, and how scattered pieces of evidence connect. These are
precisely the concerns addressed by temporal databases and stream
processing~\cite{babcock2002models,arasu2006cql,dyreson1994consensus,snodgrass2012tsql2},
yet current conversational memory systems lack analogous mechanisms for
valid-time tracking, update propagation, and versioned-state querying.
Long-term conversational QA is therefore a data-management challenge over
temporal, heterogeneous, continuously evolving interaction histories.
Failing to address it leads to hallucinated answers grounded in stale context,
missed updates, or incoherent reasoning over fragmented evidence, ultimately
undermining user trust and system reliability in deployed applications.

The core difficulty stems from a mismatch between how conversational data is
stored and how it is queried. Interactions arrive as a linear, timestamped stream
of messages, while many questions require a reconstructed state. A user may
share a travel plan, later change the destination, and eventually ask about the
current itinerary or the reasoning behind the decision. Answering accurately
depends on connecting events scattered across sessions: the initial statement,
the subsequent update, and the final outcome. A chunk-based retriever can pull a
semantically relevant but obsolete message, missing the critical update that
changes the answer. In database terms, this is a query-processing failure: the
access path induced by semantic similarity reaches topical evidence, but not
necessarily the authoritative evidence for the queried state~\cite{selinger1979access,graefe1993volcano}. The challenge therefore goes beyond relevance
estimation; it requires temporal validity, evidence links, provenance, and
conflict handling, echoing database settings where queries must return reliable
answers over incomplete or inconsistent evidence~\cite{arenas1999consistent,buneman2001and,green2007provenance,dong2013data,yin2007truth}.

Existing approaches only partially address this challenge. Dense retrieval and
retrieval-augmented generation rank independent passages by relevance rather than
state validity~\cite{karpukhin2020dense,lewis2020retrieval}. Temporal retrieval
adds time constraints, but often keeps memory units as disconnected vector
objects rather than records linked by update, contradiction, and provenance
relationships~\cite{qian2024timer4}. Long-term memory systems cache, summarize,
or organize histories into notes and hierarchies~\cite{zhong2024memorybank,xu2022beyond},
improving recall and compression while potentially blurring provenance and
obscuring the update trail needed for temporally grounded answers.
Graph-enhanced retrieval and database keyword search derive explicit structures
over textual, graph-shaped, structured, and semi-structured data~\cite{li2024graphreader,he2007blinks,chen2009keyword,tran2009top},
but these structures are usually used to locate relevant content rather than to
model query-time temporal state. As a result, current architectures can retrieve
content that appears semantically relevant without ensuring that it remains
authoritative, up to date, and consistent with later conversation history.

We argue that an ideal data model for evolving conversational QA must satisfy
three requirements. First, it must preserve a \emph{multi-granularity topology}:
events, sessions, and topics should all be queryable units, since answers often
traverse several levels. This connects conversational memory to graph and
path-query settings where answers are assembled from related records rather than
from a single tuple or document~\cite{neumann2008rdf,zou2011gstore,libkin2016querying}.
Second, it must capture \emph{state evolution}: updates and contradictions should
be modeled explicitly so obsolete facts can be discounted without being deleted,
consistent with temporal data and uncertain-lineage paradigms~\cite{dyreson1994consensus,snodgrass2012tsql2,benjelloun2006uldbs}.
Third, it must support \emph{evidence reconstruction}: query processing should
surface coherent support paths rather than isolated snippets, allowing answers to
be traced to the records and transformations that justify them~\cite{buneman2001and,green2007provenance,cheney2009provenance}. Meeting these
requirements requires reconciling temporal and graph semantics with vector
retrieval under bounded online overhead, echoing adaptive query processing under
changing evidence~\cite{avnur2000eddies,gounaris2002adaptive}.

To address these challenges, we introduce TRACE, a state-aware query processing
framework over temporal evidence graphs for conversational data. TRACE constructs
a unified evidence graph from long-running conversations, where state-bearing
events are connected by typed temporal, causal, update, and contradiction
relations, while sessions and topics serve as hierarchical structural vertices.
At query time, TRACE couples message-level retrieval with graph-guided evidence
navigation: the text channel preserves recall over raw dialogue, while the
temporal graph yields validity-aware paths that dictate how retrieved evidence
should be reconciled. This design preserves the fidelity of the original
conversational text while leveraging a structured evidence layer to reason about
current states, historical facts, and multi-hop dependencies.

This paper makes the following key contributions:
\begin{itemize}[leftmargin=*]
    \item We formalize query-time state reconstruction as a foundational bottleneck in long-term conversational QA, demonstrating analytically and empirically why relevance-centric retrieval fails when underlying facts continuously evolve.
    
 \item We introduce a temporal evidence graph that unifies events, sessions, and topics into a coherent hierarchical topology, featuring typed semantic edges, temporal validity annotations for obsolete facts, and explicit modeling primitives for updates and contradictions.

\item We design and implement TRACE, a modular offline-online architecture that integrates dense vector retrieval, hierarchical seed expansion, validity-aware path traversal, multi-signal fusion scoring, and hybrid context assembly for bounded, efficient query-time reasoning.

    \item We conduct an extensive empirical study on long-conversation QA benchmarks. Through rigorous ablations, we analyze the impact of graph evidence, structural hierarchy, and query-adaptive reasoning across temporal, multi-hop, open-domain, and preference-driven workloads.
\end{itemize}

\section{Methodology}
\label{sec:methodology}

TRACE maintains two complementary memory structures: a \emph{note memory}~$\mathcal{N}$ for lexical recall of raw conversation text, and a \emph{temporal evidence graph}~$G^\star$ for structured, state-aware reasoning over evolving facts. The note memory preserves what was said; the evidence graph tracks what currently still holds and how facts connect causally across sessions.

\begin{figure*}[t]
\centering
\includegraphics[width=\textwidth]{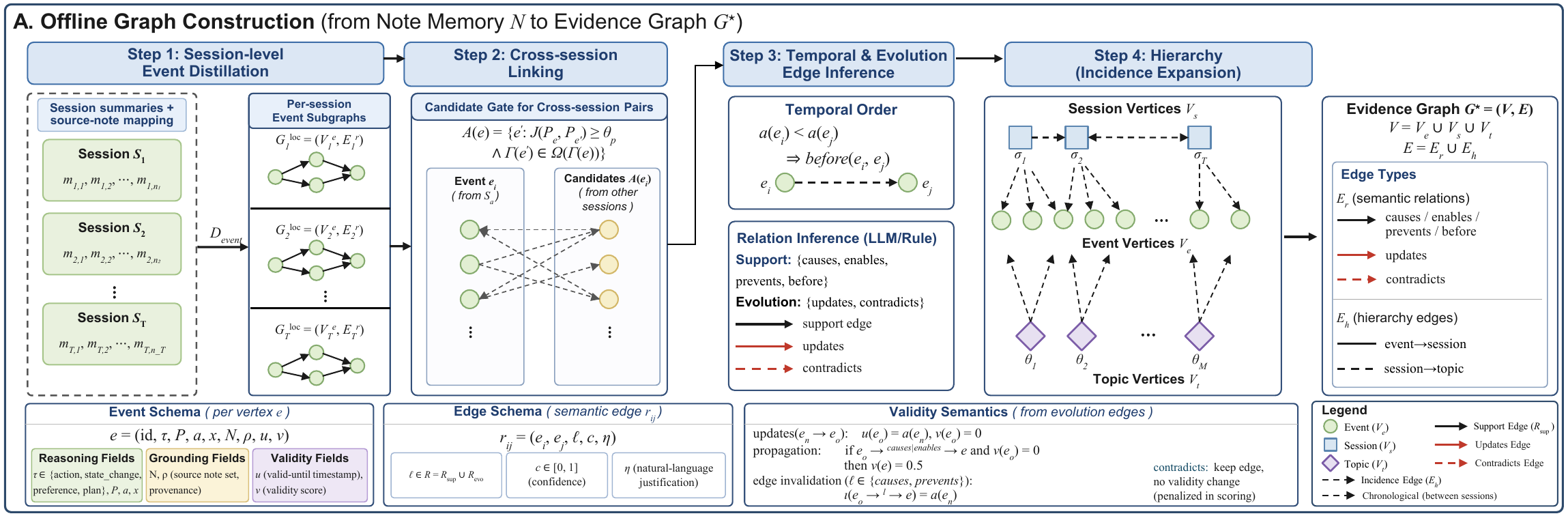}
\caption{Offline graph construction. Beginning with session-level summaries derived from note memory~$\mathcal{N}$, TRACE produces the evidence graph~$G^\star$ through four sequential steps: session-level event distillation, cross-session linking via candidate gating, temporal and evolution edge inference, and hierarchy injection.}
\label{fig:offline}
\end{figure*}

\subsection{Problem Formulation}
\label{sec:method-problem}

Let a conversation be a sequence of sessions
\begin{equation}
\mathcal{C}=\langle S_t\rangle_{t=1}^{T},\quad S_t=\langle m_{t,i}\rangle_{i=1}^{n_t},
\label{eq:conv}
\end{equation}
where each message $m_{t,i}$ carries a speaker, content, and timestamp. Given a query~$q$ posed after the conversation, the task is to produce an answer~$\hat{y}$ grounded in evidence from~$\mathcal{C}$. Unlike standard retrieval-augmented QA, evidence here must be interpreted under temporal evolution: if two retrieved passages conflict, the system must determine which reflects the current state rather than treating both as equally valid. The complete pipeline is summarized in Algorithm~\ref{alg:trace}.

\begin{algorithm}[t]
\caption{TRACE Pipeline}
\label{alg:trace}
\begin{algorithmic}[1]

\REQUIRE Conversation $\mathcal{C}=\langle S_t\rangle_{t=1}^{T}$, Memory $\mathcal{N}$, Query $q$
\ENSURE Answer $\hat{y}$

\medskip
\STATE \textbf{// Offline: Graph Construction (§\ref{sec:method-build})}
\FOR{each session $S_t$}
    \STATE $(V_t^e, E_t^r) \leftarrow D_{\mathrm{event}}(S_t)$ \hfill $\triangleright$ Eq.~(\ref{eq:distill})
\ENDFOR
\STATE Cross-session linking via candidate gate $A(e)$ \hfill $\triangleright$ Eq.~(\ref{eq:candidates})
\STATE Infer temporal \& evolution edges $\mathcal{R}_{\mathrm{sup}} \cup \mathcal{R}_{\mathrm{evo}}$
\STATE Inject hierarchy: $E_h \leftarrow \mathrm{Inc}(\mathcal{E}_S) \cup \mathrm{Inc}(\mathcal{E}_T)$ \hfill $\triangleright$ Eq.~(\ref{eq:membership})

\medskip
\STATE \textbf{// Validity Propagation (§\ref{sec:method-validity})}
\FOR{each $e_n \xrightarrow{\texttt{updates}} e_o$}
    \STATE $v(e_o) \leftarrow 0$;\; propagate to causal dependents \hfill $\triangleright$ Eq.~(\ref{eq:update-stamp}), (\ref{eq:edge-validity})
\ENDFOR

\medskip
\STATE \textbf{// Online: Query Processing (§\ref{sec:method-query})}
\STATE $V_q \leftarrow V_0(q) \cup V_{\mathrm{ent}} \cup V_{\mathrm{upd}} \cup V_{\mathrm{hier}}$ \hfill $\triangleright$ Eq.~(\ref{eq:seed0}), (\ref{eq:seed})
\STATE $\Pi \leftarrow$ bounded bidirectional traversal from $V_q$ with pruning
\FOR{each path $\pi \in \Pi$}
    \STATE $S(\pi) \leftarrow \alpha\,\mathit{NS} + \beta\,\mathit{PC} + \gamma\,\mathit{TC} + \delta\,\mathit{UV}$ \hfill $\triangleright$ Eq.~(\ref{eq:score})
\ENDFOR

\medskip
\STATE \textbf{// Context Assembly \& Generation (§\ref{sec:method-context})}
\STATE Assemble $C_{\mathrm{note}}$, $C_{\mathrm{path}}$ under budget $B_{\mathrm{total}}$ \hfill $\triangleright$ Eq.~(\ref{eq:path-budget})
\STATE Select answer plan $p_q$ by query shape
\STATE $\hat{y} \leftarrow \mathcal{A}(q,\, C_{\mathrm{note}},\, C_{\mathrm{path}},\, p_q)$ \hfill $\triangleright$ Eq.~(\ref{eq:answer})

\medskip
\RETURN $\hat{y}$

\end{algorithmic}
\end{algorithm}

TRACE maintains a composite memory state
\begin{equation}
\mathcal{M}=(\mathcal{N},\,G^\star).
\label{eq:memory}
\end{equation}
$\mathcal{N}$ preserves original conversational text and supports dense retrieval; $G^\star$ encodes how facts relate to, support, and supersede one another over time. A query is answered by combining relevant information drawn from both:
\begin{equation}
\hat{y}=\mathcal{A}(q,\,C_{\mathrm{note}},\,C_{\mathrm{path}},\,p_q).
\label{eq:answer}
\end{equation}
$C_{\mathrm{note}}$ is the lexical context from~$\mathcal{N}$ providing direct textual grounding in the original user utterances, $C_{\mathrm{path}}$ is the path-based evidence from~$G^\star$ providing relational structure and temporal validity status, and $p_q$ is a query-shape-specific answer plan. The separation is deliberate: the note memory keeps answer generation grounded in original conversational text, while the evidence graph ensures that this retrieved evidence is interpreted correctly with respect to temporal evolution and supersession relationships.

\subsection{Temporal Evidence Graph}
\label{sec:method-graph}

A flat list of extracted events suffices for single-hop factual lookup, but long-conversation queries frequently require reasoning over chains of related facts. Understanding that event~A caused event~B, which was later updated by event~C, demands explicit relational structure. Moreover, queries operate at different granularities: some ask about a specific event, others about a session's outcome, and others about a topic spanning multiple sessions. TRACE therefore represents evidence as a typed, directed graph with three vertex populations:
\begin{equation}
G=(V,E),\quad V=V_e\cup V_s\cup V_t,\quad E=E_r\cup E_h.
\label{eq:graph}
\end{equation}
\emph{Event vertices}~$V_e$ are the atomic units of factual content; \emph{session vertices}~$V_s$ preserve conversational boundaries; \emph{topic vertices}~$V_t$ group semantically related sessions and enable cross-session traversal. $E_r$ contains \emph{semantic relation edges} among events, while $E_h$ contains \emph{hierarchical membership edges} connecting events to sessions and sessions to topics.

\paragraph{Event schema}
Each event vertex encodes the minimal information needed for both reasoning and answer generation:
\begin{equation}
e=(\mathit{id},\,\tau,\,P,\,a,\,x,\,N,\,\rho,\,u,\,v).
\label{eq:event-schema}
\end{equation}
The \emph{reasoning fields} include the event type~$\tau\in\{\texttt{action}, \texttt{state\_change}, \texttt{preference}, \texttt{plan}\}$, the participant set~$P$, time anchor~$a$, and description~$x$, which together support graph traversal and temporal comparison. The \emph{grounding fields} include source note set~$N$ and provenance~$\rho$, which associate each event with the note evidence used during construction and support later retrieval and audit. The \emph{validity fields} include valid-until timestamp~$u$ and validity score~$v$: $u$ is initially unset, while $v$ is initialized as valid and updated by the validity semantics. This schema keeps the graph compact (one vertex per state-bearing fact rather than per message) while maintaining full traceability.

\paragraph{Semantic edges}
Two events connected by a semantic edge share a logical or temporal dependency:
\begin{equation}
r_{ij}=(e_i,\,e_j,\,\ell,\,c,\,\eta),\quad \ell\in\mathcal{R},\quad c\in[0,1].
\label{eq:edge-schema}
\end{equation}
Here $\ell$ is the relation label, $c$ is a confidence score, and $\eta$ is a natural-language justification supporting interpretability. The relation vocabulary is partitioned into two families with distinct downstream semantics:
\begin{equation}
\mathcal{R}=\mathcal{R}_{\mathrm{sup}}\cup\mathcal{R}_{\mathrm{evo}}.
\label{eq:relations}
\end{equation}
$\mathcal{R}_{\mathrm{sup}}=\{\texttt{causes},\texttt{enables},\texttt{prevents},\texttt{before}\}$~(the support family) captures how facts causally or temporally depend on one another within the memory graph. These edges carry information forward in time and form the backbone of multi-hop reasoning chains. The \emph{evolution family} $\mathcal{R}_{\mathrm{evo}}=\{\texttt{updates},\texttt{contradicts}\}$ captures where the conversational state changes. These edges carry the temporal validity signals needed to distinguish current from obsolete information. The distinction matters because the two families are treated differently during both validity propagation and query-time traversal: support edges form reasoning chains, while evolution edges trigger state transitions.

\begin{figure*}[t]
\centering
\includegraphics[width=\textwidth]{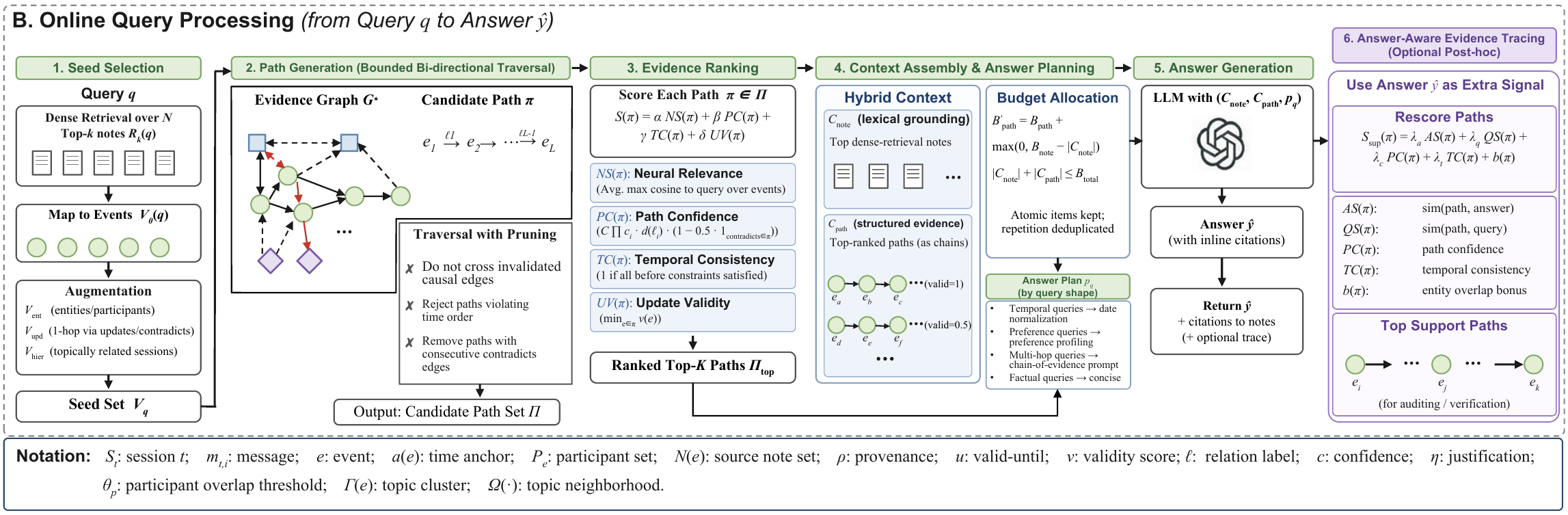}
\caption{Online query processing pipeline. Given a query~$q$, TRACE performs seed selection, bounded bidirectional path generation with pruning, evidence ranking, hybrid context assembly with answer planning, answer generation from the assembled context, and optional post-hoc evidence tracing.}
\label{fig:online}
\end{figure*}
\paragraph{Hierarchical structure}
The event--session--topic hierarchy is conceptually a nested hypergraph, but TRACE realizes it as a bipartite incidence expansion so that standard graph traversal operators can apply directly and uniformly without requiring any special-case hyperedge handling:
\begin{equation}
E_h=\mathrm{Inc}(\mathcal{E}_S)\cup\mathrm{Inc}(\mathcal{E}_T).
\label{eq:membership}
\end{equation}
$\mathrm{Inc}(S)$ inserts directed membership edges between each event $e\in S$ and the corresponding session vertex~$\sigma_S$; $\mathrm{Inc}(T)$ connects each session $\sigma\in T$ with the topic vertex~$\theta_T$. This gives TRACE a uniform traversal substrate: a query can descend from a topic to its sessions and then to specific events, or ascend from a seed event to discover related evidence elsewhere. The three granularities correspond to three common query patterns: event-level for precise factual lookup, session-level for broader contextual understanding, and topic-level for preference aggregation or longitudinal trend tracking across sessions.

\subsection{Offline Graph Construction}
\label{sec:method-build}

As illustrated in Fig.~\ref{fig:offline}, the offline stage transforms session-level textual summaries derived from the note memory~$\mathcal{N}$ into a faithful, well-connected evidence graph~$G^\star$ through four operators applied in sequence.

\paragraph{Step 1: Session-level event distillation}
The first operator extracts event vertices and intra-session edges from each session-level textual unit:
\begin{equation}
(V_t^e,\,E_t^r)=D_{\mathrm{event}}(S_t).
\label{eq:distill}
\end{equation}
The operator runs on the session-level textual summary of $S_t$ and retains source-note identifiers linking extracted events back to~$\mathcal{N}$. This granularity provides enough local context to resolve participants, temporal references, and short-range causal relations, while avoiding the fragmentation of message-level extraction and the context-window limits and long-range ambiguity of conversation-level extraction.
\paragraph{Step 2: Cross-session linking}
Local subgraphs alone cannot answer queries requiring connections across sessions (e.g., ``Did Alice follow through on the plan she mentioned last month?''). However, naively considering all cross-session event pairs is $O(|V_e|^2)$ and produces a dense graph dominated by spurious connections. TRACE introduces a candidate gate that restricts cross-session relation inference to plausible pairs:
\begin{equation}
A(e)=\{e':J(P_e,P_{e'})\geq\theta_p\;\wedge\;\Gamma(e')\in\Omega(\Gamma(e))\}.
\label{eq:candidates}
\end{equation}
$J$ is participant Jaccard overlap and $\Gamma(e)$ maps an event to its topic cluster. The participant constraint enforces a necessary condition for state evolution: one person's preference change is irrelevant to another person's plan, so events involving disjoint participants rarely need direct edges. The topic constraint prevents linking superficially similar events from unrelated threads. Together, these two lightweight filters reduce the candidate space to a sparse, semantically plausible subset without requiring expensive embedding comparisons.

\paragraph{Step 3: Temporal and evolution edge inference}
For each candidate pair passing the gate, TRACE determines the appropriate edge type. When both events have parseable time anchors, a deterministic \texttt{before} edge is added based on chronological ordering. For remaining candidates, TRACE infers whether a support or evolution relation exists. This is where the graph acquires its state-tracking capability: \texttt{updates} edges identify where newer information supersedes older facts, and \texttt{contradicts} edges flag unresolved conflicts.

\paragraph{Step 4: Hierarchy injection}
Finally, TRACE injects session and topic vertices into the graph using the incidence expansion of Eq.~\eqref{eq:membership} and adds chronological edges between adjacent session vertices. This step adds no new factual content; it purely adds navigational structure that enables multi-granularity traversal at query time.

\subsection{Validity Semantics}
\label{sec:method-validity}

The construction stage produces \texttt{updates} and \texttt{contradicts} edges, but these edges alone do not tell the query processor which events to trust. Validity semantics translate evolution edges into per-event state annotations that can be evaluated during traversal.

Each event is initialized as fully valid with $v(e)=1$, and $u(e)$ remains unconstrained until an \texttt{updates} edge is encountered. When a newer event~$e_n$ is connected to an older event~$e_o$ via an \texttt{updates} edge, TRACE performs:
\begin{equation}
u(e_o)=a(e_n),\quad v(e_o)=0.
\label{eq:update-stamp}
\end{equation}
The older event is marked superseded as of the update time. For current-state queries, $e_o$ will be deprioritized; for historical queries asking about the state before~$a(e_n)$, the event remains available because the valid-until timestamp records exactly when supersession occurred.

Validity propagates one hop further: if an event~$e$ depends on a superseded event through a \texttt{causes} or \texttt{enables} edge, its validity is reduced to $v(e)=0.5$. The intuition is that a consequence of an obsolete cause is not necessarily false (it may have already occurred), but its evidential reliability is diminished. The resulting three-valued scheme, with $1$ denoting currently valid, $0.5$ denoting partially suspect or uncertain, and $0$ denoting fully superseded, avoids the complexity of probabilistic belief propagation while still providing enough granularity for practically useful distinctions at query time.

Updates also invalidate strong causal edges incident on superseded events:
\begin{equation}
\iota(e_o\xrightarrow{\ell}e)=a(e_n),\quad \ell\in\{\texttt{causes},\texttt{prevents}\}.
\label{eq:edge-validity}
\end{equation}
Traversal at query time skips edges with a set invalidation timestamp, preventing reasoning chains from passing through obsolete causal links. Importantly, \texttt{contradicts} edges behave differently: they do not modify validity scores but are retained as explicit conflict signals and penalized during path scoring. This asymmetry reflects a semantic distinction. An update resolves a state change where the newer version is authoritative, while a contradiction records disagreement without resolution. Silently invalidating one side of a contradiction would inject an unsupported judgment.

\subsection{Online Query Processing}
\label{sec:method-query}

With the evidence graph fully constructed and validity annotations in place, TRACE answers queries by combining lexical recall from~$\mathcal{N}$ with structured traversal over~$G^\star$, as illustrated in the overview shown in Fig.~\ref{fig:online}.

\paragraph{Seed selection}
Bridging from an unstructured query to the structured graph is the first challenge. TRACE uses dense retrieval over~$\mathcal{N}$ to obtain a top-$k$ note set~$R_k(q)$, then maps these to event vertices:
\begin{equation}
V_0(q)=\{e:N(e)\cap R_k(q)\neq\emptyset\}.
\label{eq:seed0}
\end{equation}
Vector similarity alone has well-known blind spots, including lexical mismatch, failure to retrieve updated versions, and inability to recover cross-session evidence with different vocabulary. TRACE compensates by augmenting $V_0$:
\begin{equation}
V_q=V_0\cup V_{\mathrm{ent}}\cup V_{\mathrm{upd}}\cup V_{\mathrm{hier}}.
\label{eq:seed}
\end{equation}
$V_{\mathrm{ent}}$ comes from entity and participant lookup, $V_{\mathrm{upd}}$ adds one-hop neighbors along evolution edges to ensure that both old and new versions are properly considered, and $V_{\mathrm{hier}}$ adds events from topically related sessions via the hierarchy. Each expansion family targets a distinct retrieval failure mode.

\paragraph{Path generation}
From~$V_q$, TRACE performs bounded bidirectional traversal to generate candidate evidence paths:
\begin{equation}
\pi=\langle e_1,\ell_1,e_2,\ldots,\ell_{L-1},e_L\rangle.
\label{eq:path}
\end{equation}
Each path represents a chain of evidence connecting multiple facts through explicit logical relations. Three hard pruning rules apply during traversal: invalidated causal edges are never crossed because they represent broken reasoning links; paths violating parseable timestamps are rejected to prevent backward causation; and paths with consecutive \texttt{contradicts} edges are removed because chaining two unresolved conflicts produces no reliable inference.

\paragraph{Evidence ranking}
Each surviving path is then scored by a combination of four signals:

\begin{equation}
S(\pi)=\alpha\,\mathit{NS}(\pi)+\beta\,\mathit{PC}(\pi)+\gamma\,\mathit{TC}(\pi)+\delta\,\mathit{UV}(\pi).
\label{eq:score}
\end{equation}
The default weights $(\alpha,\beta,\gamma,\delta)=(0.4,0.3,0.15,0.15)$ reflect priority ordering: relevance first, structural confidence second, temporal signals for refinement.

\emph{Neural relevance} averages per-event similarity to the query:
\begin{equation}
\mathit{NS}(\pi)=L^{-1}\sum_{e\in\pi}\max_{n\in N(e)}\cos(\phi(q),\phi(n)).
\label{eq:ns}
\end{equation}
The average rather than maximum penalizes paths containing one relevant event padded with irrelevant intermediate hops, thereby encouraging compact and focused evidence chains.

\emph{Path confidence} captures structural reliability:
\begin{equation}
\mathit{PC}(\pi)=\prod_{i=1}^{L-1}c_i\,d(\ell_i)\cdot\bigl(1-0.5\,\mathbf{1}\{\texttt{contradicts}\in\pi\}\bigr).
\label{eq:pc}
\end{equation}
The product form means a single weak link degrades the entire chain, which is appropriate because a reasoning chain is only as strong as its weakest individual step. The relation discount~$d(\ell_i)$ assigns lower values to weaker or less informative relation types, and the contradiction penalty halves confidence for paths that traverse unresolved or ambiguous conflicts.

\emph{Temporal consistency} verifies chronological constraints:
\begin{equation}
\mathit{TC}(\pi)=\mathbf{1}\bigl\{a(e_i)\leq a(e_j)\ \text{for every}\ e_i\xrightarrow{\texttt{before}}e_j\in\pi\bigr\}.
\label{eq:tc}
\end{equation}

\emph{Update validity} reflects reliance on superseded information:
\begin{equation}
\mathit{UV}(\pi)=\min_{e\in\pi}v(e).
\label{eq:uv}
\end{equation}
The minimum operator ensures even a single superseded event makes staleness visible. Paths through obsolete events are not forbidden, as they may be needed for historical queries, but are ranked lower by default.

\subsection{Context Assembly and Answer Planning}
\label{sec:method-context}

The evidence graph produces structured paths, but the final answer must ultimately be generated by a language model consuming text. TRACE assembles a hybrid context with two sections: $C_{\mathrm{note}}$ contains top dense-retrieval results providing verbatim grounding, and $C_{\mathrm{path}}$ contains top-ranked graph paths rendered as compact support chains with explicit relation labels and validity markers.

Budget is allocated between sections:
\begin{equation}
B'_{\mathrm{path}}=B_{\mathrm{path}}+\max(0,\,B_{\mathrm{note}}-|C_{\mathrm{note}}|),
\label{eq:path-budget}
\end{equation}
\begin{equation}
|C_{\mathrm{note}}|+|C_{\mathrm{path}}|\leq B_{\mathrm{total}}.
\label{eq:total-budget}
\end{equation}
When dense retrieval returns few relevant notes (common for cross-session queries where relevant text is distributed thinly across the history), the freed budget is reallocated to additional evidence paths providing the structural context that the notes alone lack. Items are kept as atomic units during truncation, and repeated identifiers are deduplicated.

TRACE selects an answer plan~$p_q$ based on the detected query shape. Temporal queries receive date-normalization instructions directing the model to compare timestamps before asserting currentness. Preference queries receive a preference-profiling plan that aggregates evolving statements over time. Multi-hop queries receive chain-of-evidence prompts closely following path structure. Factual queries receive a concise plan prioritizing directness and brevity. This decouples the what-is-available view (evidence graph) from the how-to-reason instruction (answer plan), allowing the same evidence to be consumed under different reasoning constraints.

\subsection{Answer-Aware Evidence Tracing}
\label{sec:method-support}

After answer generation, users or downstream systems may need to verify provenance. TRACE provides an optional post-hoc tracing stage using the generated answer as an additional retrieval signal:
\begin{equation}
\resizebox{.91\linewidth}{!}{$\displaystyle S_{\mathrm{sup}}(\pi)=\lambda_a\,\mathit{AS}(\pi)+\lambda_q\,\mathit{QS}(\pi)+\lambda_c\,\mathit{PC}(\pi)+\lambda_t\,\mathit{TC}(\pi)+b(\pi).$}
\label{eq:support}
\end{equation}
$\mathit{AS}$ measures similarity between the path and the generated answer to ensure relevance to what was actually said, $\mathit{QS}$ measures similarity to the original query to ensure the trace addresses the question, $\mathit{PC}$ and $\mathit{TC}$ retain their definitions from evidence ranking, and $b(\pi)$ is a bonus for paths passing through events whose descriptions overlap with answer entities. Candidate paths with no lexical overlap with the query--answer pair are filtered as clearly irrelevant. This stage does not revise the answer; it operates purely as a post-hoc explanation mechanism, exposing the evidence chain for auditing whether the answer is faithfully grounded.

\section{Experiments}
\label{sec:experiment}

We conduct experiments to answer the following questions:
\begin{itemize}[leftmargin=*]
\item \textbf{RQ1:} Does TRACE outperform existing memory systems on long-term conversational QA across diverse question types and dataset scales?
\item \textbf{RQ2:} Is TRACE robust across LLM backbones?
\item \textbf{RQ3:} How does each component of TRACE (evidence graph, hierarchy, query-shape-aware reasoning) contribute to the overall performance?
\item \textbf{RQ4:} How does TRACE handle superseded facts, contradictions, and chain revisions that stress update-aware reasoning?
\item \textbf{RQ5:} What are the construction costs and online efficiency characteristics of TRACE's evidence graph?
\item \textbf{RQ6:} Does the evidence graph provide interpretable and faithful reasoning traces?
\end{itemize}

\subsection{Experimental Setup}
\label{sec:setup}

\paragraph{Datasets}
We evaluate TRACE on two long-term conversational QA benchmarks
spanning moderate-length and haystack-scale dialogue.
\textbf{LoCoMo}~\cite{maharana2024evaluating} provides a publicly released
$10$-conversation subset averaging $24$K tokens per dialogue
($\sim$$9$K over the full $50$-conversation set) with $1{,}540$
non-adversarial questions.
\textbf{LongMemEval\textsubscript{S}}~\cite{wu2024longmemeval} contains $500$
questions over dialogue histories averaging $\sim$$115$K tokens, of
which $470$ are non-adversarial and span six evidence-based
question types; the original \emph{abstention} (false-premise) type
is excluded. Following prior work, adversarial questions are excluded
on both benchmarks.

\paragraph{Stress test benchmark}
LoCoMo and LongMemEval evaluate long-term memory broadly, but
the slices that exercise temporal update reasoning are heavily
diluted by single-hop and simple recall questions that do not
require any form of state-aware retrieval. To specifically probe
how a memory system handles superseded facts, contradictions, and
chain revisions in isolation (RQ4), we construct a dedicated
update-heavy benchmark called the \emph{Stress Test}, which is
entirely derived from LoCoMo conversations.
We mine LoCoMo session pairs for typed-edge candidates carrying
update or contradiction signals, yielding $772$ raw candidates
across four type labels: \texttt{chain\_modification},
\texttt{fact\_override},
\texttt{contradiction\_resolution}, and
\texttt{temporal\_correction}. From this pool, we apply
\emph{coverage-based representative selection} to retain $100$
candidates that maximize coverage over (sample, type,
signal-pattern) tuples, and then manually re-author each against
the LoCoMo session summaries and dialog text. This yields the
final $59$-item benchmark in natural descriptive style (mean $8.8$
words per reference), preserving both type coverage ($22/18/15/4$
across the four update types) and difficulty coverage (medium $28$,
hard $16$, easy $15$). Edge-direction ambiguities are resolved by
treating the LoCoMo session index as the chronology authority,
where a larger index corresponds to a later point in time.

Each item contains the following fields: the question, the
\texttt{current\_valid\_answer}, the \texttt{outdated\_answer}
(i.e., the answer that was correct under an earlier session state
but has since been superseded), an \texttt{evidence\_path} that
records the chain of events and edges connecting the current state
to its predecessors, the source LoCoMo sample index, and a
difficulty label. The \texttt{outdated\_answer} field plays a key
role in our evaluation protocol: it allows the temporal-aware judge
to separately measure \texttt{outdated\_leakage} in addition to
\texttt{valid\_acc}, so that a system returning a previously correct
but now superseded answer is penalized rather than rewarded—even
though that answer was valid at some earlier point in the
conversation. We note that our selection procedure operates as an
engineered heuristic over discrete
$(\text{sample}, \text{type})$ buckets rather than a formal coreset
algorithm; to avoid overstating the theoretical guarantee, we use
the term \emph{representative selection} throughout the paper.
\paragraph{Baselines}
We compare TRACE against seven representative systems whose design
positions are reviewed in Sec.~\ref{sec:relatedwork}.
\textbf{Full-Context} concatenates the full dialogue history into
the LLM context and serves as a non-retrieval reference.
\textbf{RAG} indexes each turn as an independent note and retrieves
the top-$10$ turn-level notes by dense
similarity~\cite{karpukhin2020dense,lewis2020retrieval}, with no
LLM-based memory evolution.
\textbf{LangMem}~\cite{langmem2026} runs in a backbone-specific variant:
under GPT-4o-mini we mount LangMem's memory tools on a LangGraph
\texttt{create\_react\_agent} hot-path agent; under DeepSeek-V4-Flash
(which does not autonomously emit memory calls) we substitute
LangMem's program-driven \texttt{create\_memory\_store\_manager}
variant.
\textbf{Mem0}~\cite{chhikara2025mem0} retrieves the top-$30$
semantically similar facts per speaker at query time.
\textbf{Nemori}~\cite{nan2025nemori} is reproduced from the official
release on our matched cohort, using the same backbone LLM and
embedding model as all other systems.
\textbf{Zep}~\cite{rasmussen2025zep} is reproduced at session
granularity with top-$20$ edge and node retrieval and reciprocal
rank-fusion reranking.
\textbf{A-Mem}~\cite{xu2026mem} shares the same per-turn note schema
and top-$10$ retrieval as RAG, and additionally invokes the backbone
LLM at ingestion time to evolve cross-note links on LoCoMo; this
evolution step is disabled on LongMemEval\textsubscript{S}, where
the 199K-turn cache renders full evolution computationally infeasible. TRACE inherits
A-Mem's per-turn note ingestion as its bottom layer
(Sec.~\ref{sec:method-build}); we therefore designate A-Mem as the
primary point of comparison and report $\Delta$ rows against it in
Tables~\ref{tab:locomo_main} and~\ref{tab:lme_main}.
TRACE plus the seven baselines are evaluated end-to-end on LoCoMo
($n{=}1540$) under both backbones; on LongMemEval\textsubscript{S}
($n{=}470$) we compare TRACE against A-Mem and Nemori, with
Full-Context as a non-retrieval reference.

\begin{table*}[t!]
\caption{Main results on LoCoMo by backbone and question category.
}
\label{tab:locomo_main}
\centering
\resizebox{\textwidth}{!}{%
\begin{tabular}{l|ccc|ccc|ccc|ccc|ccc}
\toprule
 & \multicolumn{3}{c|}{\textbf{Temporal}}
 & \multicolumn{3}{c|}{\textbf{Open Domain}}
 & \multicolumn{3}{c|}{\textbf{Multi-Hop}}
 & \multicolumn{3}{c|}{\textbf{Single-Hop}}
 & \multicolumn{3}{c}{\textbf{Overall}} \\
\textbf{Method}
 & LLM & F1 & BLEU & LLM & F1 & BLEU & LLM & F1 & BLEU & LLM & F1 & BLEU & LLM & F1 & BLEU \\
\midrule
\multicolumn{16}{l}{\textit{Backbone: GPT-4o-mini}} \\
\midrule
Full-Context
 & 0.435 & 0.379 & 0.330 & 0.441 & 0.170 & 0.134 & 0.706 & 0.274 & 0.211 & 0.881 & 0.450 & 0.333 & 0.728 & 0.385 & 0.297 \\
\midrule
RAG
 & 0.462 & 0.370 & 0.316 & 0.289 & 0.103 & 0.084 & 0.396 & 0.221 & 0.156 & 0.547 & 0.363 & 0.304 & 0.486 & 0.322 & 0.266 \\
LangMem
 & 0.215 & 0.262 & 0.222 & 0.435 & \textbf{0.249} & \textbf{0.198} & \underline{0.537} & \underline{0.333} & \underline{0.254} & 0.583 & 0.366 & 0.296 & 0.488 & 0.331 & 0.267 \\
Zep
 & 0.331 & 0.390 & 0.343 & 0.344 & 0.188 & 0.155 & 0.480 & 0.314 & 0.213 & 0.436 & 0.324 & 0.276 & 0.416 & 0.327 & 0.271 \\
Mem0
 & 0.288 & 0.360 & 0.332 & 0.444 & \underline{0.230} & \underline{0.180} & 0.497 & 0.323 & 0.228 & 0.624 & 0.394 & 0.313 & 0.519 & 0.364 & 0.293 \\
Nemori
 & \underline{0.507} & \underline{0.482} & \textbf{0.409} & \underline{0.476} & 0.205 & 0.151 & 0.510 & 0.284 & 0.190 & \textbf{0.688} & \underline{0.459} & \underline{0.366} & \underline{0.604} & \underline{0.416} & \underline{0.329} \\
A-Mem
 & 0.494 & 0.413 & 0.357 & 0.270 & 0.113 & 0.092 & 0.503 & 0.254 & 0.187 & 0.589 & 0.379 & 0.314 & 0.533 & 0.347 & 0.286 \\
\rowcolor{gray!15}
\textbf{TRACE (ours)}
 & \textbf{0.693} & \textbf{0.535} & \underline{0.388}
 & \textbf{0.583} & 0.133 & 0.104
 & \textbf{0.599} & \textbf{0.357} & \textbf{0.280}
 & \underline{0.679} & \textbf{0.489} & \textbf{0.439}
 & \textbf{0.661} & \textbf{0.452} & \textbf{0.379} \\
\hspace{1em}$\Delta$ vs Nemori
 & +18.6 & +5.3 & -2.1 & +10.7 & -7.2 & -4.7 & +8.9 & +7.3 & +9.0 & -0.9 & +3.0 & +7.3 & +5.7 & +3.6 & +5.0 \\
\hspace{1em}$\Delta$ vs A-Mem
 & +19.9 & +12.2 & +3.1 & +31.4 & +2.0 & +1.2 & +9.7 & +10.3 & +9.3 & +9.1 & +11.0 & +12.5 & +12.8 & +10.6 & +9.3 \\
\midrule
\multicolumn{16}{l}{\textit{Backbone: DeepSeek-V4-Flash}} \\
\midrule
Full-Context
 & 0.524 & 0.343 & 0.282 & 0.472 & 0.213 & 0.172 & 0.775 & 0.446 & 0.323 & 0.884 & 0.658 & 0.593 & 0.764 & 0.526 & 0.452 \\
\midrule
RAG
 & 0.435 & 0.301 & 0.260 & 0.385 & 0.202 & 0.157 & 0.400 & 0.270 & 0.165 & 0.574 & 0.413 & 0.363 & 0.502 & 0.350 & 0.293 \\
LangMem$^{\dagger}$
 & 0.351 & 0.356 & 0.273 & 0.447 & 0.170 & 0.140 & 0.411 & 0.243 & 0.170 & 0.519 & 0.296 & 0.240 & 0.460 & 0.291 & 0.228 \\
Zep
 & \underline{0.660} & \underline{0.528} & \textbf{0.450} & 0.406 & \underline{0.241} & 0.186 & 0.606 & 0.371 & 0.251 & 0.552 & 0.399 & 0.339 & 0.575 & 0.411 & 0.337 \\
Mem0
 & 0.577 & 0.501 & 0.382 & 0.451 & 0.222 & \underline{0.199} & 0.597 & 0.364 & 0.254 & 0.629 & 0.398 & 0.321 & 0.601 & 0.402 & 0.314 \\
Nemori
 & 0.621 & 0.522 & \underline{0.430} & 0.500 & \textbf{0.284} & \textbf{0.207} & \textbf{0.693} & \textbf{0.408} & \underline{0.286} & \underline{0.730} & 0.491 & 0.401 & \underline{0.686} & \underline{0.470} & \underline{0.374} \\
A-Mem
 & 0.462 & 0.298 & 0.249 & 0.420 & 0.224 & 0.178 & 0.610 & 0.368 & 0.259 & 0.686 & \underline{0.498} & \underline{0.443} & 0.609 & 0.416 & 0.352 \\
\rowcolor{gray!15}
\textbf{TRACE (ours)}
 & \textbf{0.723} & \textbf{0.559} & 0.425
 & \textbf{0.601} & 0.116 & 0.089
 & \underline{0.671} & \underline{0.407} & \textbf{0.336}
 & \textbf{0.758} & \textbf{0.562} & \textbf{0.512}
 & \textbf{0.725} & \textbf{0.505} & \textbf{0.435} \\
\hspace{1em}$\Delta$ vs Nemori
 & +10.2 & +3.7 & -0.5 & +10.1 & -16.8 & -11.8 & -2.2 & -0.1 & +5.0 & +2.8 & +7.1 & +11.1 & +3.9 & +3.5 & +6.1 \\
\hspace{1em}$\Delta$ vs A-Mem
 & +26.1 & +26.1 & +17.6 & +18.1 & -10.8 & -8.9 & +6.2 & +3.9 & +7.7 & +7.2 & +6.4 & +6.9 & +11.6 & +8.9 & +8.3 \\
\bottomrule
\end{tabular}%
}
\quad\\\quad\\
Bold marks the best memory-augmented system within each backbone;
\underline{underline} marks the runner-up.
$\Delta$ rows report TRACE minus A-Mem and TRACE minus Nemori in
percentage points (pp).
$^{\dagger}$see Sec.~\ref{sec:setup} for the LangMem variant on
DeepSeek-V4-Flash.
\end{table*}
\paragraph{Backbones}
To probe cross-backbone robustness (RQ2), we evaluate every system with
two LLMs of distinct families: \texttt{gpt-4o-mini} (OpenAI, accessed
through the OpenRouter API) and \texttt{deepseek-v4-flash} (DeepSeek,
accessed through the official DeepSeek API). The same backbone is used
for both memory construction and answer generation within each
setting.

\paragraph{Metrics}
Following the LLM-as-a-judge paradigm~\cite{zheng2023judging}, our
primary metric on both benchmarks is the binary LoCoMo grading
protocol of Mem0~\cite{chhikara2025mem0}, as also adopted by
Nemori~\cite{nan2025nemori}, in which a separate judge LLM scores each
generated answer against the reference as a binary correctness
verdict. The judge is \texttt{gpt-4o-mini} at temperature $0$ with
JSON output, run three times independently; we report the mean score
across runs. On LoCoMo, we additionally report token-level F1 and
BLEU-1 of the original evaluation protocol~\cite{maharana2024evaluating}
to align with prior reports; on LongMemEval\textsubscript{S}, we report
F1 only and omit BLEU-1, since references are verbose free-form
sentences for which n-gram overlap is dominated by stylistic surface
form rather than answer correctness. For the Stress Test (RQ4), we
report \texttt{valid\_acc} (fraction of correct current-state answers),
\texttt{outdated\_leakage} (fraction returning the superseded answer),
and \texttt{strict net} (= valid\_acc $-$ outdated\_leak) under a
strict-aware judging protocol.

\paragraph{Implementation details}
All retrieval-based systems use the \texttt{all-MiniLM-L6-v2} sentence
encoder~\cite{reimers2019sentence,wang2020minilm}. TRACE uses one
configuration across both benchmarks; LongMemEval\textsubscript{S}
applies haystack-scale adapters (chronological session chunking instead of LLM topic clustering, with update
detection disabled because LongMemEval has one fact per topic,
no within-session contradictions, and prohibitive pairwise checks at scale) that leave retrieval and
scoring unchanged. Online answer generation culminates in one final LLM call per
question on a $10$K-token hybrid context (notes plus causal evidence).
Memory caches and graph indices are built once per (system, backbone,
dataset) and reused across reruns.

\subsection{Main Results (RQ1, RQ2)}
\label{sec:main}

\begin{table*}[t!]
\caption{Main results on LongMemEval\textsubscript{S} by backbone and
question type.
}
\label{tab:lme_main}
\centering
\resizebox{\textwidth}{!}{%
\begin{tabular}{l|cc|cc|cc|>{\columncolor{gray!15}}c>{\columncolor{gray!15}}c|cc|cc}
\toprule
 & \multicolumn{2}{c|}{\textbf{Full-Context}}
 & \multicolumn{2}{c|}{\textbf{A-Mem}}
 & \multicolumn{2}{c|}{\textbf{Nemori}}
 & \multicolumn{2}{c|}{\textbf{TRACE (ours)}}
 & \multicolumn{2}{c|}{\textbf{$\Delta$ vs Nemori}}
 & \multicolumn{2}{c}{\textbf{$\Delta$ vs A-Mem}} \\
\textbf{Question type}
 & LLM & F1 & LLM & F1 & LLM & F1 & LLM & F1 & LLM & F1 & LLM & F1 \\
\midrule
\multicolumn{13}{l}{\textit{Backbone: GPT-4o-mini}} \\
\midrule
single-session-user (64)
 & 0.781 & 0.176 & 0.781 & 0.600 & \textbf{0.933} & \textbf{0.768} & \underline{0.797} & \underline{0.641} & -13.6 & -12.7 & +1.6 & +4.1 \\
single-session-preference (30)
 & 0.489 & 0.112 & \underline{0.367} & \underline{0.140} & 0.328 & 0.113 & \textbf{0.900} & \textbf{0.451} & +57.2 & +33.8 & +53.3 & +31.1 \\
single-session-assistant (56)
 & 0.798 & 0.166 & \underline{0.839} & \underline{0.681} & 0.604 & 0.505 & \textbf{0.964} & \textbf{0.755} & +36.0 & +25.0 & +12.5 & +7.4 \\
multi-session (121)
 & 0.433 & 0.035 & 0.339 & 0.235 & \textbf{0.613} & \underline{0.254} & \underline{0.402} & \textbf{0.343} & -21.1 & +8.9 & +6.3 & +10.8 \\
temporal-reasoning (127)
 & 0.402 & 0.127 & 0.399 & \textbf{0.355} & \underline{0.476} & \underline{0.344} & \textbf{0.606} & 0.329 & +13.0 & -1.5 & +20.7 & -2.6 \\
knowledge-update (72)
 & 0.736 & 0.122 & 0.681 & 0.387 & \underline{0.756} & \underline{0.446} & \textbf{0.773} & \textbf{0.479} & +1.7 & +3.3 & +9.2 & +9.2 \\
\textbf{Overall (470)}
 & 0.565 & 0.113 & 0.529 & 0.387 & \underline{0.622} & \underline{0.399} & \textbf{0.667} & \textbf{0.457} & \textbf{+4.5} & \textbf{+5.8} & \textbf{+13.8} & \textbf{+7.0} \\
\midrule
\multicolumn{13}{l}{\textit{Backbone: DeepSeek-V4-Flash}} \\
\midrule
single-session-user (64)
 & 0.984 & 0.184 & \underline{0.922} & \underline{0.767} & 0.667 & 0.604 & \textbf{0.938} & \textbf{0.790} & +27.1 & +18.6 & +1.6 & +2.3 \\
single-session-preference (30)
 & 0.900 & 0.151 & 0.356 & 0.101 & \underline{0.374} & \underline{0.114} & \textbf{0.833} & \textbf{0.413} & +45.9 & +29.9 & +47.7 & +31.2 \\
single-session-assistant (56)
 & 0.982 & 0.285 & \textbf{1.000} & \textbf{0.828} & 0.319 & 0.276 & \underline{0.964} & \underline{0.798} & +64.5 & +52.2 & -3.6 & -3.0 \\
multi-session (121)
 & 0.678 & 0.058 & 0.408 & 0.254 & \underline{0.423} & \underline{0.386} & \textbf{0.534} & \textbf{0.445} & +11.1 & +5.9 & +12.6 & +19.1 \\
temporal-reasoning (127)
 & 0.732 & 0.135 & 0.449 & \underline{0.372} & \textbf{0.748} & \textbf{0.464} & \underline{0.656} & 0.270 & -9.2 & -19.4 & +20.7 & -10.2 \\
knowledge-update (72)
 & 0.819 & 0.121 & 0.616 & 0.420 & \underline{0.647} & \underline{0.437} & \textbf{0.796} & \textbf{0.566} & +14.9 & +12.9 & +18.0 & +14.6 \\
\textbf{Overall (470)}
 & 0.806 & 0.139 & \underline{0.588} & \underline{0.440} & 0.563 & 0.414 & \textbf{0.733} & \textbf{0.503} & \textbf{+17.0} & \textbf{+8.9} & \textbf{+14.5} & \textbf{+6.3} \\
\bottomrule
\end{tabular}%
}
\quad\\\quad\\
Bold marks the best memory-augmented system within each row;
\underline{underline} marks the runner-up; Full-Context is excluded
from the ranking pool as a non-retrieval reference.
$\Delta$ columns report TRACE minus A-Mem and TRACE minus Nemori
in percentage points (pp).

\end{table*}

\paragraph{Overall effectiveness (RQ1)}
Table~\ref{tab:locomo_main} reports results on LoCoMo across four
question categories. Under GPT-4o-mini, TRACE achieves the highest
overall LLM judge score ($0.661$), F1 ($0.452$), and BLEU-1
($0.379$), outperforming the strongest retrieval baselines Nemori
and A-Mem by $+5.7$ and $+12.8$\,pp on LLM score, respectively.
The gains are most pronounced on Temporal questions
($+18.6$\,pp vs.\ Nemori, $+19.9$\,pp vs.\ A-Mem in LLM score)
and Multi-Hop questions ($+8.9$/$+9.7$\,pp), both of which
require reasoning over cross-session evidence chains, precisely
the capability targeted by TRACE's path-based evidence ranking. On Single-Hop questions TRACE
remains competitive with Nemori ($-0.9$\,pp LLM) while exceeding
it on F1 ($+3.0$) and BLEU ($+7.3$), indicating that the graph
layer does not harm simple retrieval performance.

On the Open-Domain category, TRACE leads on LLM 
score ($+10.7$\,pp vs.\ Nemori) but trails on F1 and BLEU. These
questions are broad conversational prompts with verbose
multi-sentence references; TRACE's concise answers receive
high semantic judgments but lower n-gram overlap. The
same pattern appears across both backbones and benchmarks, indicating
that TRACE's advantage is clearest under
correctness-oriented metrics
rather than surface-overlap
metrics.

Table~\ref{tab:lme_main} extends the evaluation to the
haystack-scale LongMemEval\textsubscript{S} ($\sim$115K tokens per
dialogue, 470 questions). Under GPT-4o-mini, TRACE achieves
$0.667$ LLM and $0.457$ F1 overall, surpassing Nemori by
$+4.5$/$+5.8$\,pp and A-Mem by $+13.8$/$+7.0$\,pp. On the
single-session-preference type, which requires aggregation of evolving
user preferences, TRACE improves over both baselines by over
$+50$\,pp (LLM), validating the design decision to include
preference-profiling answer plans. On single-session-assistant
questions, which probe retrieval of assistant-generated content,
TRACE's $0.964$ LLM score demonstrates near-ceiling performance.
The multi-session category, the most demanding cross-session
reasoning type, sees a $+6.3$\,pp LLM and $+10.8$\,pp F1
improvement over A-Mem.
TRACE's temporal-reasoning results on LongMemEval\textsubscript{S}
reveal an interesting trade-off: TRACE achieves the highest LLM
score ($0.606$) under GPT-4o-mini but trails Nemori on F1
($0.329$ vs.\ $0.344$). Manual inspection shows that
LongMemEval's temporal-reasoning reference answers contain precise date
strings; TRACE's evidence paths sometimes retrieve the correct
temporal relation but express it through relative phrasing, which
penalizes exact token overlap while still preserving semantic correctness.

\paragraph{Cross-backbone robustness (RQ2)}
Both tables demonstrate that TRACE's overall rankings are preserved
under DeepSeek-V4-Flash, a backbone from a different model family
with a distinct tokenizer and instruction-following profile.
On LoCoMo (Table~\ref{tab:locomo_main}), TRACE leads overall
($0.725$ LLM, $0.505$ F1, $0.435$ BLEU) by $+3.9$/$+11.6$\,pp
over Nemori/A-Mem on LLM score. On
LongMemEval\textsubscript{S} (Table~\ref{tab:lme_main}), TRACE
achieves $0.733$ LLM and $0.503$ F1 overall,
with the largest single-category gain again on
single-session-preference ($+45.9$\,pp vs.\ Nemori,
$+47.7$\,pp vs.\ A-Mem). The absolute performance on 
DeepSeek is generally higher than GPT-4o-mini, consistent with the stronger Full-Context ceiling, yet TRACE's relative advantage over baselines remains stable across both backbones. This consistency confirms that the performance gains stem from the evidence graph structure and query-processing pipeline rather than from backbone-specific prompt engineering.

The one notable exception is the single-session-assistant type on
DeepSeek, where A-Mem achieves a perfect $1.000$ LLM score,
suggesting the task is nearly solved by strong dense retrieval
within a single session on this backbone. TRACE's $0.964$ on the
same slice is not meaningfully different from ceiling, and the
$-3.6$\,pp gap reflects one or two borderline judgments rather
than a systematic failure. A second exception is temporal-reasoning
on DeepSeek, where Nemori's LLM score ($0.748$) exceeds TRACE
($0.656$). DeepSeek's stronger instruction-following ability
amplifies Nemori's timeline-oriented prompting strategy on this
category; however, TRACE's overall advantage across all six
categories remains substantial ($+17.0$\,pp LLM overall).

\subsection{Ablation Study (RQ3)}
\label{sec:ablation}

\begin{table}[t]
\caption{Compact overall ablation summary on GPT-4o-mini}
\label{tab:ablation_summary}
\centering
\resizebox{0.95\columnwidth}{!}{%
\begin{tabular}{l|ccc|cc}
\toprule
 & \multicolumn{3}{c|}{\textbf{LoCoMo}}
 & \multicolumn{2}{c}{\textbf{LongMemEval\textsubscript{S}}} \\
\textbf{Variant} & LLM & F1 & BLEU & LLM & F1 \\
\midrule
\rowcolor{gray!15}
\textbf{R0 TRACE} & \textbf{0.661} & \textbf{0.452} & \textbf{0.379} & \textbf{0.667} & \textbf{0.457} \\
R1 w/o Topic & 0.660 & 0.449 & 0.374 & 0.652 & 0.450 \\
R2 w/o Hier & 0.644 & 0.446 & 0.374 & 0.650 & 0.455 \\
R3 w/o graph & 0.599 & 0.403 & 0.332 & 0.609 & 0.438 \\
R4 w/o L3 & 0.609 & 0.369 & 0.313 & 0.550 & 0.385 \\
R5 \emph{A-Mem} & \emph{0.533} & \emph{0.347} & \emph{0.286} & \emph{0.529} & \emph{0.387} \\
\bottomrule
\end{tabular}%
}
\end{table}

\begin{figure*}[t]
\centering
\includegraphics[width=\textwidth]{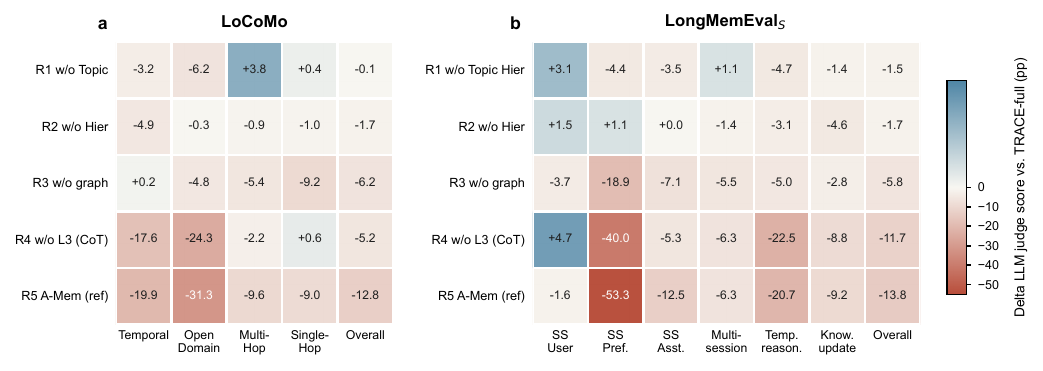}
\caption{Ablation impact across benchmarks. Each cell shows the LLM-score change in percentage points relative to TRACE for an ablated variant
on (a) LoCoMo and (b) LongMemEval\textsubscript{S}. Negative values indicate performance loss after removing a component.}
\vspace{-3mm}
\label{fig:ablation-impact}
\end{figure*}

\paragraph{Component contributions}
Table~\ref{tab:ablation_summary} reports overall ablation metrics,
while Fig.~\ref{fig:ablation-impact} summarizes category-level LLM
changes. Each row removes one component from the full system (R0);
row R5 reproduces the A-Mem baseline for reference.

\emph{Topic hierarchy (R1: w/o Topic)} Removing topic vertices
reduces overall LoCoMo LLM score by only $0.1$\,pp but
LongMemEval\textsubscript{S} LLM by $1.5$\,pp. The small LoCoMo
effect is expected because that benchmark's 10 conversations are
short enough that session-level edges already provide sufficient
cross-context links. On the haystack-scale
LongMemEval\textsubscript{S}, topic vertices serve as
navigational shortcuts across hundreds of sessions, and their
removal affects temporal-reasoning ($-4.7$\,pp) and
single-session-assistant ($-3.5$\,pp) most. This validates the
multi-granularity topology design principle: at scale, topic-level grouping
becomes a necessary structural scaffold for efficient traversal.

\emph{Full hierarchy (R2: w/o Hier)} Removing both topic and
session vertices reduces LoCoMo overall LLM to $0.644$
($-1.7$\,pp) and LongMemEval\textsubscript{S} LLM to $0.650$
($-1.7$\,pp). The fact that R2 remains well above the A-Mem
reference (R5) by $+11.1$/$+12.1$\,pp overall confirms that the
event-level graph, together with the same query-shape-aware
retrieval and answer-planning stack, provides substantial value. The hierarchy adds a
consistent but moderate additional gain on top of the
event-semantic foundation.

\emph{Graph evidence (R3: w/o graph)} When the entire graph
layer is removed and the system falls back to note-only retrieval
with query-shape-aware prompting, LoCoMo overall LLM drops to
$0.599$ ($-6.2$\,pp) and LongMemEval\textsubscript{S} LLM to
$0.609$ ($-5.8$\,pp). The gap concentrates on categories that
require relational reasoning: Multi-Hop on LoCoMo ($-5.4$\,pp),
Single-Hop ($-9.2$\,pp), and single-session-preference on
LongMemEval\textsubscript{S} ($-18.9$\,pp). This demonstrates
that path-based evidence provides information that flat retrieval
cannot recover, even when the same underlying notes are available,
because the graph encodes \emph{how} facts relate rather than
merely \emph{which} facts exist. The R3 result also establishes
that the query-shape planner alone (without graph evidence) still
outperforms A-Mem by $+6.6$/$+8.0$\,pp overall, confirming that
the planner provides independent value through better reasoning
instructions.

\emph{Query-shape-aware reasoning (R4: w/o L3 CoT)} Replacing the query-shape-specific answer plan with a generic prompt
produces the largest single-component drop on LongMemEval\textsubscript{S}
and one of the largest drops on LoCoMo: $-5.2$\,pp overall
LLM on LoCoMo and $-11.7$\,pp on LongMemEval\textsubscript{S}.
On LoCoMo, the Temporal category is most affected ($-17.6$\,pp
LLM), consistent with the design rationale that temporal queries
need explicit date-normalization reasoning. On
LongMemEval\textsubscript{S}, the single-session-preference type
collapses from $0.900$ to $0.500$ ($-40.0$\,pp), confirming that
the preference-profiling plan is high-impact for aggregating
evolving statements into a coherent current-state answer. Notably,
R4 still outperforms R5 (A-Mem) overall by $+7.6$/$+2.1$\,pp,
indicating that graph evidence contributes value even without the
tailored plan, but the two components are strongly complementary:
good evidence without good reasoning instructions yields only
partial improvement, and the converse also holds.
\paragraph{Summary}
The ablation reveals a layered contribution structure: query-shape-aware reasoning (L3) and graph evidence provide the
two main gains with benchmark- and question-dependent balance:
L3 supports plan-aware temporal/preference reasoning, while graph
paths recover evidence missed by flat retrieval for factual and
multi-hop questions; the hierarchical
structure provides a modest but consistent gain by enabling
multi-granularity navigation. No single component accounts for
the full improvement over baselines, confirming that TRACE's
effectiveness emerges from the interaction of its architectural
layers rather than any single mechanism.

\subsection{Update-Heavy Slice Results (RQ4)}
\label{sec:update-slice}

\begin{table}[t!]
\caption{Update-tracking on the Stress test and the C3 update-aware seed ablation}
\label{tab:stress_c3}
\centering
\resizebox{0.95\columnwidth}{!}{%
\begin{tabular}{l|c|ccc}
\toprule
 & \textbf{LoCoMo full}
 & \multicolumn{3}{c}{\textbf{Stress test ($n{=}59$)}} \\
\cmidrule(lr){3-5}
\textbf{System / Variant}
 & LLM ($n{=}1540$) & valid\_acc & outdated\_leak & strict net \\
\midrule
Full-Context
 & $0.728$ & $0.542$ & $0.076$ & $+0.466$ \\
Mem0
 & $0.519 \pm 0.0031$ & $0.339$ & $\mathbf{0.051}$ & $\underline{+0.288}$ \\
Nemori
 & $\underline{0.604 \pm 0.0010}$ & $0.322$ & $0.102$ & $+0.220$ \\
A-Mem
 & $0.533 \pm 0.0040$ & $\underline{0.356}$ & $0.085$ & $+0.271$ \\
\rowcolor{gray!15}
\textbf{TRACE (ours)}
 & $\mathbf{0.661 \pm 0.0004}$ & $\mathbf{0.390}$ & $\underline{0.076}$ & $\mathbf{+0.314}$ \\
\quad w/o update-aware seed
 & $0.652 \pm 0.0017$ & $0.305$ & $0.093$ & $+0.212$ \\
\midrule
\multicolumn{5}{l}{\textit{$\Delta$ TRACE vs.\ baselines:}} \\
\quad vs.\ Nemori
 & $+5.70$ & $+6.8$ & $-2.6$ & $+9.4$ \\
\quad vs.\ A-Mem
 & $+12.81$ & $+3.4$ & $-0.8$ & $+4.2$ \\
\multicolumn{5}{l}{\textit{$\Delta$ C3 ablation (pp):}} \\
\quad Overall
& $+0.93$ & $+8.47$ & $-1.69$ & $+10.17$ \\
\quad Open-Domain only
 & $+7.29$ & --- & --- & --- \\
\midrule
\multicolumn{5}{l}{\textit{Evolution-OFF diagnostic:}} \\
\quad A-Mem
 & --- & $0.186$ & $0.110$ & $+0.076$ \\
\quad TRACE
 & --- & $0.254$ & $0.102$ & $+0.153$ \\
\bottomrule
\end{tabular}%
}
\end{table}

\begin{table}[t!]
\caption{Graph statistics, build cost, and online efficiency across benchmarks and backbones}
\label{tab:graph_summary}
\centering
\resizebox{0.95\columnwidth}{!}{%
\begin{tabular}{l|cc|cc}
\toprule
 & \multicolumn{2}{c|}{\textbf{LoCoMo (10 samples)}}
 & \multicolumn{2}{c}{\textbf{LongMemEval\textsubscript{S} (limit\,500)}} \\
\textbf{Metric}
 & GPT-4o-mini & DeepSeek & GPT-4o-mini & DeepSeek \\
\midrule
\multicolumn{5}{l}{\textit{Corpus}} \\
\hspace{1em}Notes
 & 5{,}882   & 5{,}882   & 199{,}509 & 199{,}509 \\
\hspace{1em}Sessions
 & 272       & 272       & 19{,}195  & 19{,}195 \\
\hspace{1em}Topics
 & 50        & 49        & 384       & 384 \\
\midrule
\multicolumn{5}{l}{\textit{Graph}} \\
\hspace{1em}Events
 & 2{,}288   & 2{,}537   & 128{,}376 & 151{,}988 \\
\hspace{1em}Edges (total)
 & 7{,}156   & 8{,}151   & 398{,}025 & 454{,}991 \\
\hspace{1em}Update-aware edges
 & 185       & 362       & 1         & 8 \\
\hspace{1em}Structural / Semantic (\%)
 & 74.6 / 25.4 & 72.6 / 27.4 & 78.9 / 21.1 & 79.5 / 20.5 \\
\multicolumn{5}{l}{\hspace{1em}\textit{Hierarchy fanout (avg)}} \\
\hspace{2em}Events / Session
 & 8.4       & 9.3       & 6.7       & 7.9 \\
\hspace{2em}Sessions / Topic
 & 5.4       & 5.6       & 50.0      & 50.0 \\
\midrule
\multicolumn{5}{l}{\textit{Construction}} \\
\hspace{1em}Build wall-clock
 & 4.73~h    & 5.85~h    & 38.03~h   & 44.55~h \\
\hspace{1em}Time per event
 & 7.45~s    & 8.3~s     & 0.92~s    & 1.05~s \\
\hspace{1em}Storage (graph JSON)
 & 6.0~MB    & 6.6~MB    & 273~MB    & 318~MB \\
\midrule
\multicolumn{5}{l}{\textit{Online cost}} \\
\hspace{1em}Tokens / Q\ \ (TRACE)
 & 3{,}719   & 4{,}391   & 6{,}948   & 6{,}058 \\
\hspace{1em}Tokens / Q\ \ (A-Mem)
 & 6{,}642   & 8{,}345   & 2{,}313   & 2{,}644 \\
\hspace{1em}Tokens / Q\ \ (Full-Context)
 & 15{,}931  & 21{,}535  & 98{,}547  & 104{,}206 \\
\midrule
\multicolumn{5}{l}{\textit{Online latency (median ms / Q)}} \\
\hspace{1em}TRACE
 & 947       & 728       & 1{,}171   & 1{,}025 \\
\hspace{1em}A-Mem
 & 703       & 564       & 663       & 345 \\
\hspace{1em}Full-Context
 & 718       & 372       & 6{,}424   & 5{,}079 \\
\bottomrule
\end{tabular}%
}
\end{table}

\paragraph{Stress test performance}
Table~\ref{tab:stress_c3} evaluates how well each system handles
superseded facts, contradictions, and chain revisions on the
purpose-built 59-item Stress Test. TRACE
achieves the highest valid accuracy ($0.390$) and strict net score
($+0.314$) among memory-augmented systems, surpassing Nemori by $+9.4$\,pp and A-Mem by
$+4.2$\,pp on strict net. Crucially, TRACE's outdated leakage
($0.076$) matches the Full-Context reference and is lower than both
Nemori ($0.102$) and A-Mem ($0.085$). This validates the
update-validity semantics: by
marking superseded events with $v(e){=}0$ and deprioritizing paths
that traverse them via the $\mathit{UV}(\pi)$ scoring term
(Eq.~\ref{eq:uv}), TRACE avoids surfacing stale facts as if they
were current.

The comparison with Full-Context is also instructive. Full-Context
achieves the highest valid accuracy ($0.542$) by having access to
the entire conversation, but its outdated leakage ($0.076$) is
identical to TRACE's. This suggests that TRACE's retrieval pipeline
successfully identifies the same temporal resolution cues that are
available in the full context, while using $5{\times}$ fewer
tokens. The gap in valid accuracy ($0.542$ vs.\ $0.390$) indicates
that some update chains in the Stress Test require evidence
distributed so broadly across the conversation that bounded path
traversal does not fully recover all supporting passages, a known
limitation of any retrieval-based approach relative to full-context
processing.

\paragraph{Update-aware seed ablation (C3)}
Disabling the update-aware seed expansion ($V_{\mathrm{upd}}$ in
Eq.~\ref{eq:seed}) causes the Stress Test strict net to drop from
$+0.314$ to $+0.212$ ($-10.17$\,pp after accounting for both
valid\_acc and leak changes). The valid accuracy drops by
$-8.47$\,pp and outdated leakage increases by $+1.69$\,pp,
confirming that the expansion along evolution edges is essential
for retrieving the most recent version of a revised fact. Without
this expansion, the system relies solely on dense similarity to
find the current state; when the current version uses different
vocabulary from the query (a common pattern in preference changes),
the older, more lexically similar version is retrieved instead.
On the broader LoCoMo benchmark, the ablation reduces overall LLM
score by $-0.93$\,pp, with the effect concentrated in the
Open-Domain sub-category ($-7.29$\,pp), a category where user
preferences frequently evolve and where the update-aware seed
ensures the current preference is retrieved alongside the
historical one.

\begin{table*}[t!]
\caption{Three-layer interpretability evaluation of TRACE}
\label{tab:interp_stack}
\centering
\resizebox{\textwidth}{!}{%
\renewcommand{\arraystretch}{1.15}
\begin{tabular}{@{}p{1.9cm} p{2.7cm} p{4.3cm} p{6.4cm}@{}}
\toprule
\textbf{Layer} & \textbf{Probe ($n$)} & \textbf{TRACE headline} & \textbf{Defensive evidence / cross-layer link} \\
\midrule
Graph &
Edge Audit ($100$) &
$94.0\%$ factual / $89.0\%$ type-label accuracy &
The $11$\,pp type-label error concentrates in the temporal class
(\textsc{temporal\_before} on same-timestamp events). \\
\addlinespace
Pre-reasoning &
Cue Glanceability ($22$) &
$81.8\%$ current support; $3.73/5$ glanceability &
$+63.6$\,pp vs.\ semantic-only baseline ($18.2\%$); misleading rate
$22.7\%$ vs.\ $54.6\%$. \\
\addlinespace
Post-reasoning &
Path Faithfulness ($145$) &
Per-cat: SH $65.0\%$, MH $45.0\%$, OD $36.0\%$, \textbf{Tem $22.5\%$};
overall $42.5\%$ &
Independent rule-based corroborator: rule-clean $65.4\%$ vs.\
rule-flagged $37.8\%$, $\boldsymbol{+27.6}$\,pp. Cat~$2$ weakness
$\boldsymbol{90.3\%}$ propagated from the graph layer; cat~$1$
$+5.9$\,pp $\approx$ flat (negative control). \\
\bottomrule
\end{tabular}%
}
The temporal-class type-label errors observed at the graph layer reappear at the
explanation layer as the dominant driver of cat~$2$ path-faithfulness failure
($90.3\%$ propagation), confirmed by an independent rule-based corroborator's
$+27.6$\,pp clean-vs-flagged gap.
\end{table*}

\paragraph{Cache-controlled diagnostic}
To test whether the Stress Test results merely inherit A-Mem's evolved
note cache, we run a two-by-two cache diagnostic using the default
evolution cache and a UUID-remapped no-evolution cache. The no-evolution
cache strips A-Mem's ingestion-time note links and evolved metadata while
preserving graph-to-note lookup compatibility for TRACE. Removing note
evolution reduces A-Mem's strict net from $+0.271$ to $+0.076$
($-19.5$\,pp), while TRACE drops from $+0.314$ to $+0.153$
($-16.1$\,pp). On the matched no-evolution cache, TRACE still retains a
direction-consistent $+7.7$\,pp advantage over A-Mem. As a cross-arm
diagnostic, comparing the paper-default TRACE setting against A-Mem
without ingestion-time evolution increases the gap to $+23.7$\,pp. These results
indicate that A-Mem's stress performance is heavily supported by
ingestion-time evolution metadata, whereas TRACE retains an independent
graph-side signal even after the metadata has been stripped away.

\subsection{Graph Construction and Efficiency Analysis (RQ5)}
\label{sec:cost}

% \begin{figure*}[t]
% \centering
% \includegraphics[width=0.95\textwidth]{Figure/figure4.pdf}
% \caption{Representative update-aware evidence traces. TRACE separates stale memories from current support through typed temporal and update
% relations, enabling current-state answers while preserving historical evidence}
% \label{fig:update-trace}
% \end{figure*}

\paragraph{Graph statistics}
Table~\ref{tab:graph_summary} characterizes the evidence graph at
both benchmark scales. On LoCoMo (10 conversations, 5{,}882 notes),
TRACE distills approximately 2{,}300--2{,}500 events with
7{,}100--8{,}200 edges, yielding an average fanout of ${\sim}3.1$
edges per event. On LongMemEval\textsubscript{S} (19{,}195 sessions,
199{,}509 notes), the graph scales to 128--152K events with
398--455K edges. The structural-to-semantic edge ratio is
approximately 75/25 at both scales, reflecting the dominance of
membership and session-ordering edges over the rarer semantic
relations. Here structural edges refer to membership and
session-level \texttt{temporal\_before} edges, while semantic edges
include \texttt{causes}, \texttt{enables}, \texttt{prevents},
event-level \texttt{temporal\_before}, \texttt{updates}, and
\texttt{contradicts}. Update-aware edges (\texttt{updates} $+$
\texttt{contradicts}) number 185--362 on LoCoMo but only 1--8
on LongMemEval\textsubscript{S}; this is expected because the latter
benchmark's conversations rarely revisit the same topic within a
single dialogue sequence, meaning within-conversation fact evolution
is sparse. Note that Sessions/Topic on
LongMemEval\textsubscript{S} equals the fixed chunk size of $50$,
as LLM topic clustering is bypassed by chronological windows at the
$19$K-session scale. Nonetheless, the temporal hierarchy and
session-ordering structure still enable multi-session reasoning on
LongMemEval\textsubscript{S}, as demonstrated by the multi-session
category gains in Table~\ref{tab:lme_main}.

\paragraph{Construction cost}
Build time is 4.7--5.9~hours on LoCoMo (7.5--8.3\,s per event)
and 38--45~hours on LongMemEval\textsubscript{S} (0.9--1.1\,s
per event). The order-of-magnitude speedup in per-event time at
larger scale is attributable to the candidate gating mechanism
(Eq.~\ref{eq:candidates}): with more events, the participant and
topic filters prune a proportionally larger candidate space,
amortizing the cost of LLM inference for edge classification over
more frequent rejection of trivial pairs. Storage is modest: 6\,MB
for LoCoMo and 273--318\,MB for the full
LongMemEval\textsubscript{S} graph (JSON serialization). These
numbers indicate that the offline construction is a one-time
investment whose cost is dominated by LLM calls rather than
storage or memory, and that the candidate gate successfully
controls the quadratic blowup that would otherwise make
cross-session linking intractable at scale.

\paragraph{Online cost and latency}
TRACE's median online latency is 947/728\,ms per question on
LoCoMo and 1{,}171/1{,}025\,ms on LongMemEval\textsubscript{S}
(GPT-4o-mini/DeepSeek). The latency overhead compared to A-Mem
(703/564\,ms on LoCoMo) comes from graph traversal and path
scoring; however, TRACE remains well within interactive response
bounds ($<$1.2\,s). On LongMemEval\textsubscript{S}, Full-Context
requires 6.4/5.1\,s and over 98--104K tokens per question.
TRACE reduces this to 6{,}948/6{,}058 tokens, a
$14{\times}$--$17{\times}$ reduction, while achieving higher
accuracy (Table~\ref{tab:lme_main}). The efficiency gain derives
directly from the bounded path generation: rather than feeding the entire
conversation to the LLM, TRACE supplies only the top-scored paths
and notes within the 10K-token budget.
The token comparison also highlights an important design trade-off
relative to A-Mem.

On LoCoMo, TRACE uses fewer tokens than A-Mem
(3{,}719 vs.\ 6{,}642 under GPT-4o-mini) because note
deduplication against graph paths compresses redundancy. On
LongMemEval\textsubscript{S}, TRACE uses more tokens (6{,}948
vs.\ 2{,}313) because the path budget is allocated to
cross-session evidence that A-Mem's flat retrieval does not attempt
to surface. This selective budget expansion directly enables the
multi-session and temporal-reasoning gains observed in
Table~\ref{tab:lme_main}, demonstrating that the additional tokens
carry high-value structural information rather than undifferentiated
context padding.

\subsection{Interpretability Analysis (RQ6)}
\label{sec:interpret}

Table~\ref{tab:interp_stack} assesses TRACE's interpretability at
three layers: graph fidelity, pre-reasoning cue support, and
post-reasoning explanation faithfulness.

\emph{Graph layer: Edge Audit} On a random sample of 100 edges,
human annotators confirm $94.0\%$ factual accuracy and $89.0\%$
type-label accuracy. The $11$\,pp type-label error concentrates in
the temporal class (\texttt{temporal\_before} on same-timestamp
events), where session-level fallback ordering occasionally produces
incorrect direction assignments.

\emph{Pre-reasoning layer: Cue Glanceability} In a 22-question pilot,
$81.8\%$ of TRACE's rendered cues contain the information needed to
answer correctly, vs.\ $18.2\%$ for a similarity-only baseline
($+63.6$\,pp). The misleading rate is $22.7\%$ vs.\ $54.6\%$, and
the average glanceability rating is $3.73/5$.

\emph{Post-reasoning layer: Path Faithfulness} On 145 annotated
triples, overall path faithfulness is $42.5\%$ (Wilson 95\% CI
$[35.0, 50.9]$): Single-Hop $65.0\%$, Multi-Hop $45.0\%$,
Open-Domain $36.0\%$, Temporal $22.5\%$. The temporal weakness traces
to graph-layer type-label errors: $90.3\%$ of category-2 failures
involve a misclassified temporal edge, confirmed by an independent
rule-based corroborator ($p{=}0.010$, $+27.6$\,pp gap).

These results confirm two interpretability properties: the typed-edge
vocabulary makes state-transition reasons human-readable without
inspecting raw text, and the co-presence of stale and current nodes
preserves full revision history for auditing. The temporal edge
imprecision at construction time propagates through path ranking into
the generated answer, indicating that improving time-anchor resolution
(e.g., turn-level ordering) would yield compounding gains across all
three layers. The high fidelity on non-temporal edges (${\sim}97\%$)
confirms that the typed-edge schema provides a reliable interpretability
foundation for the majority of question types.

% \section{Discussion and Limitations}
% \label{sec:discussion}

% Discussion — TBD
% Section 7: Discussion / Limitations.
%   - C1 (architectural novelty) vs C2 (empirical primary driver) tension framing
%   - Update-mechanism behavior on the Stress test slice
%   - LongMemEval scale-conditional structural mechanisms (skip_evolution +
%     skip_update_detection caveats)
%   - Full-Context cross-backbone Pareto: TRACE +10.2pt over FC on
%     4o-mini LongMemEval; FC +7.4pt over TRACE on DS LongMemEval;
%     cost-vs-quality framing across backbones
%   - Path faithfulness pilot soft ceiling

\section{Related Work}
\label{sec:relatedwork}

\paragraph{Long-term memory and temporal validity}
Systems such as A-Mem~\cite{xu2026mem},
Mem0~\cite{chhikara2025mem0}, Nemori~\cite{nan2025nemori},
Zep~\cite{rasmussen2025zep}, and LangMem~\cite{langmem2026}
address what should be persisted across interactions, storing
notes, entities, summaries, graph facts, or interaction traces.
These systems make memory persistent and searchable, but most stored
objects are exposed to the downstream model as flat retrieved artifacts
rather than as a structured state-reconstruction problem that accounts for
how facts evolve over a conversation history. On the temporal side,
databases formalize valid time and transaction time
semantics~\cite{dyreson1994consensus,snodgrass2012tsql2}, while
uncertainty and lineage models attach confidence and derivation context
to individual records~\cite{benjelloun2006uldbs,cheney2009provenance}. Temporal KG
QA methods retrieve and rerank facts under time constraints~\cite{qian2024timer4},
but they typically begin from an already well-structured knowledge graph
with explicitly recorded temporal annotations, an assumption that does not
hold for raw conversational data.
TRACE adopts memory-note ingestion as its lexical layer, but promotes the
reasoning object to a temporal evidence graph derived from raw dialogue
sessions, where validity annotations, update edges, and contradiction
edges collectively determine how past statements should affect current answers,
bridging the gap between memory persistence and temporally coherent reasoning.

\paragraph{Graph-based retrieval, path reasoning, and evidence provenance}
Graph-based retrieval systems organize unstructured corpora into explicit
structures for search and reasoning. GraphReader~\cite{li2024graphreader}
agentically walks an entity graph for long-document QA. Database keyword
search and graph-query systems provide the query-processing counterpart,
using schema graphs, tuple trees, ranked answer graphs, RDF exploration,
and subgraph matching to connect relevant
records~\cite{agrawal2002dbxplorer,bhalotia2002keyword,hristidis2002discover,he2007blinks,chen2009keyword,tran2009top,neumann2008rdf,zou2011gstore,libkin2016querying}.
Provenance further ensures that answers remain traceable to supporting
records~\cite{buneman2001and,green2007provenance,cheney2009provenance},
enabling downstream consumers to verify the derivation chain from source
data to final output.
These approaches improve reasoning and inspectability, but primarily
target corpus organization or graph query answering over relatively stable
knowledge bases where facts do not routinely supersede one another.
TRACE differs in treating graph paths as temporal evidence objects for dialogue
state: a path may contain superseded, supporting, or conflicting events,
and its temporal consistency directly affects how the final answer is assembled
and which nodes are promoted or demoted during generation.

\paragraph{Adaptive query processing over evolving conversational evidence}
The query-processing view in TRACE is closest to database work that
adapts execution behavior according to data and query conditions.
Classical optimizers select access paths and plans from
statistics~\cite{selinger1979access,graefe1993volcano}, adaptive query
processing revises execution at runtime in response to changing data
distributions~\cite{avnur2000eddies,gounaris2002adaptive,deshpande2007adaptive},
and stream systems support standing queries over continuously arriving
inputs with bounded memory and latency~\cite{babcock2002models,chen2000niagaracq,madden2002continuously,arasu2006cql}.
View maintenance studies how derived results should remain current when
base data changes~\cite{blakeley1986efficiently,palpanas2002incremental},
a concern that closely parallels keeping conversational answers consistent
as new turns introduce updates or retractions.
TRACE couples this adaptive principle to temporal evidence semantics,
selecting query-shape-specific plans and expanding seeds through
validity-aware graph paths to adapt to both question complexity and
evidence dynamics.

\section{Conclusion}
\label{sec:conclusion}

We presented TRACE, a state-aware query processing framework
that models conversations as a hierarchical evidence graph with
typed causal, temporal, and update edges and graded validity
annotations. At query time, it combines vector retrieval with
graph-guided path search and query-shape-aware answer planning.
Experiments on two benchmarks across two backbones show
consistent gains over existing memory systems, particularly on
temporal, multi-hop, and preference questions. On a purpose-built
Stress Test, TRACE matches the Full-Context outdated leakage
while using $5{\times}$ fewer tokens. Ablations confirm that
graph evidence, hierarchy, and query-shape reasoning each
contribute complementary value.
Future work includes incremental graph construction for
deployed assistants, turn-level time-anchor resolution to reduce
temporal edge errors, and extension to multi-agent collaboration
traces.

\bibliographystyle{IEEEtran}
\bibliography{IEEE}
\end{document}